\documentclass{ecai}
\usepackage{graphicx}
\usepackage{latexsym}

\usepackage{amsmath,array}
\usepackage{kantlipsum}
\usepackage{diagbox}
\usepackage{amsfonts}
\usepackage{tabularx}
\usepackage{wrapfig}
\usepackage[export]{adjustbox}

\usepackage{multirow}
\usepackage{makecell}
\usepackage{mathtools}
\usepackage{chronology}
\usepackage{paralist}

\usepackage{color}
\usepackage{xcolor}
\usepackage{color, colortbl}

\usepackage[pagebackref=false,breaklinks=true,colorlinks,bookmarks=false]{hyperref}

\begin{document}

\begin{frontmatter}

\title{GAN-generated Faces Detection: A Survey and New Perspectives}

\author[1,3]{\fnms{Xin}~\snm{Wang}
}
\author[1]{\fnms{Hui}~\snm{Guo}
}
\author[2]{\fnms{Shu}~\snm{Hu}
}
\author[3]{\fnms{Ming-Ching}~\snm{Chang}
} 
\author[1]{\fnms{Siwei}~\snm{Lyu}
} 

\address[1]{ University at Buffalo, SUNY, USA. 
~~\{xwang264, hguo8, siweilyu\}@buffalo.edu }
\address[2]{  Indiana University–Purdue University Indianapolis, USA. ~~hu968@purdue.edu}
\address[3]{ University at Albany, SUNY, USA. ~~\{xwang56, mchang2\}@albany.edu}

\begin{abstract}
 Generative Adversarial Networks (GAN) have led to the generation of very realistic face images,  which have been used in fake social media accounts and other disinformation matters that can generate profound impacts. Therefore, the corresponding GAN-face detection techniques are under active development that can examine and expose such fake faces. In this work, we aim to provide a comprehensive review of recent progress in GAN-face detection. We focus on methods that can detect face images that are generated or synthesized from GAN models. We classify the existing detection works into four categories: (1) deep learning-based, (2) physical-based, (3) physiological-based methods, and (4) evaluation and comparison against human visual performance. For each category, we summarize the key ideas and connect them with method implementations. We also discuss open problems and suggest future research directions.
\end{abstract}

\end{frontmatter}

\section{Introduction}

\begin{figure}[t]
\centerline{
  \includegraphics[width=0.48\textwidth]{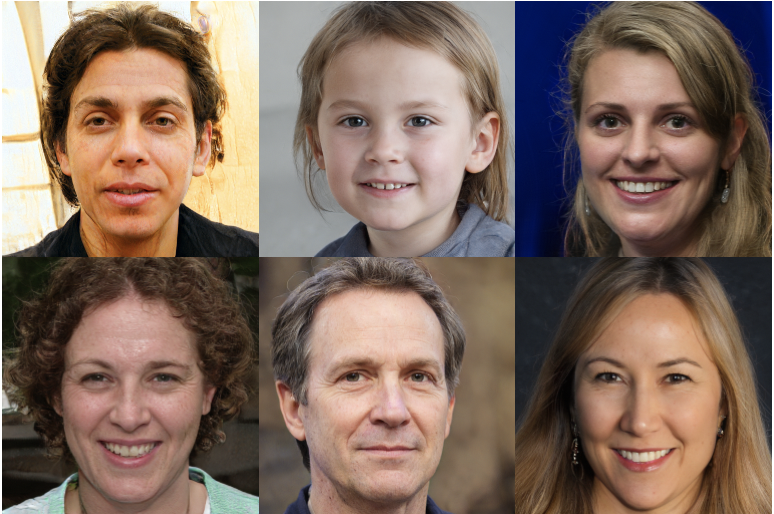}
  \vspace{-0.1em}
}
\caption{\em Examples of GAN faces generated by StyleGAN \cite{karras2019style} (\textbf{left}), StyleGAN2 \cite{karras2020analyzing} (\textbf{middle}), and StyleGAN3 \cite{karras2021alias} (\textbf{right}).
}    
\vspace{-0.1em}
\label{fig:gan-example}
\end{figure}

The development of Generative Adversarial Networks (GANs) \cite{goodfellow2014generative} enables generating high-realistic human faces images that are visually difficult to discern from real ones \cite{karras2017progressive,karras2019style,karras2020analyzing}, some examples\footnote{\url{https://thispersondoesnotexist.com}} are shown in Figure~\ref{fig:gan-example}.
GAN-generated faces (GAN-faces) can be easily used in creating fake social media accounts \cite{news1,news2,news3,news4} for malicious purposes that cause significant social concerns.
For example, a high school student created a fake candidate by using a GAN-generated face in a voting event that tricked Twitter into obtaining a coveted blue checkmark, thereby verifying the authenticity of the fake candidacy \cite{news1}. 
This fake candidate passing verification could set up donation channels to absorb public funds, which not only damages property-related laws but also diminishes election integrity. 
Furthermore, 
the fake social media accounts used GAN-faces as profile images which also generate serious negative social impacts \cite{news2,mundra2023exposing}.
For example, these fake accounts can be associated with numerous high-level executives within a company, and if they were to post comments about the company's financial situation, it could cause significant disruption in the stock market. 
This is because the false information they spread can mislead investors and cause them to make incorrect financial decisions, leading to significant losses.


Automatic detection of GAN-faces is of emerging need \cite{nightingale2022synthetic}, so numerous detection approaches have been developed to combat the malicious use of GAN-faces. However, effective GAN-face detection is still a complex and difficult problem, which typically suffers from two major challenges. First, an accurate and flexible GAN-face detection method should be able to expose the large variation of GAN-face images synthesized or generated from numerous GAN models, while remaining robust to adversarial attacks. Secondly, the decision process and the detection result should be \textbf{explainable to human users}, especially for non-AI experts, instead of only fitting to specific datasets via complex deep networks. 

\begin{figure*}[t]
\centering
\includegraphics[width=1\textwidth]{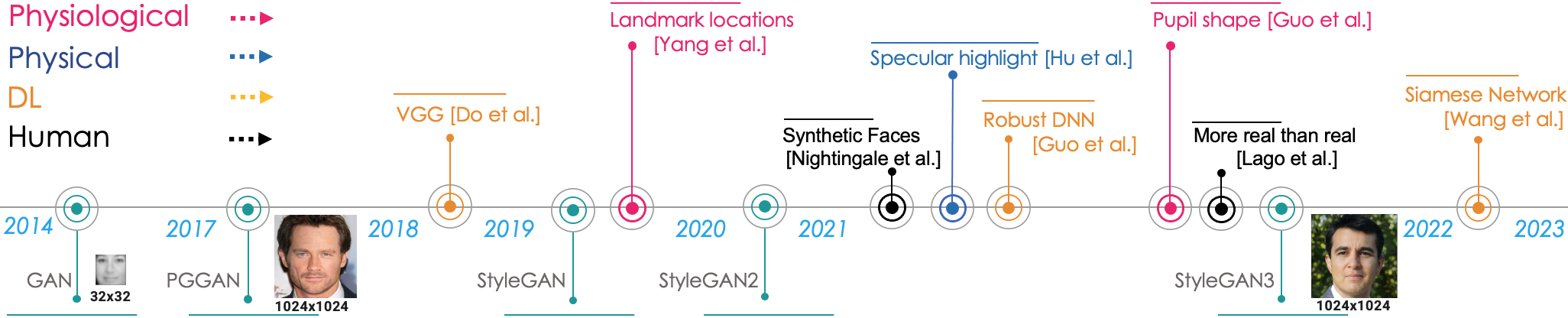}
\vspace{-1em}
\caption{\em A brief chronology for GAN-faces generation and detection works. \textbf{Generation:} The initial GAN model is proposed in 2014 and can only generate 32$\times$32 faces.  After 2017 the series of StyleGAN models can generate high-realistic faces that are hard to spot from human eyes. 
\textbf{Detection:} The earliest detection techniques are mainly based on DNN in 2018. Due to their limitation of performance and interpretability, methods based on physical and physiological cues are developed in 2019 $\sim$ 2021. 
Since StyleGAN2 generated faces are very difficult to discern from human eyes, human visual performance on GAN-generated faces is under active investigation since 2021. The listed methods represent milestones and breakthroughs in the chronology. See $\S$~\ref{method} for complete survey. }
\label{fig_timeline}
\end{figure*}



In this paper, we focus the scope of the survey on the detection of \textcolor{black}{\textbf{GAN-based entire face synthesis}}\footnote{It is different from face manipulation, which only manipulates the existing face images, instead of generation from scratch.}, which was a significant milestone of the Artificial Intelligence Generated Content (AIGC) \cite{cao2023comprehensive}. 
And this is the right time to deal with the massive use of generative AI \cite{epstein2023art}.
We start our survey by chronologically summarizing major GAN-face generation milestones ($\S$~\ref{sec:GAN:face:gen}) as well as GAN-face detection methods with highlights of important breakthroughs along with in Figure~\ref{fig_timeline}. Early GAN-face detection methods are mainly Deep Learning (DL)-based methods \cite{do2018forensics,marra2019gans}, {\em etc.}; see $\S$~\ref{sec:DL:methods}. Although they achieve promising performance in practice, it is difficult to explain the under-taking mechanism or decisions being made.  


The above limitations are overcome by approaches reasoning upon physical cues ($\S$~\ref{sec:physical}) or physiological cues ($\S$~\ref{sec:physiological}) that are explainable in nature. Recent works in this category distinguish GAN-faces by exploring the inadequacy of the GAN synthesis models in representing human faces and their corresponding relations in the physical world~\cite{yang2019exposinggan,li2018detection,matern2019exploiting}. For example, \cite{hu2021exposing} inspect the inconsistency of the corneal specular highlights between the two eyes. However, these methods work under strict assumptions such as frontal portrait faces or a clearly visible reflector in the eyes.
To eliminate these limitations and explore more robust models, \cite{guo2021eyes} introduce a physiological-based method by examining pupil shape inconsistencies.
As the human eye provides the optics and photoreception for the visual system, the pupil should generally be circular on the eye surface or appear to be elliptical in the image when viewed with an orientation. The key idea is that physiological inconsistency artifacts between the eyes ({\em e.g.} difference from comparing the boundary of pupil shapes) can be identified to distinguish GAN-faces from real faces.






An important aspect of the GAN-face detection in contrast to other AI problems (such as image classification) is that {\em 
human performance for GAN-face detection is much worse than AI algorithmic methods}. As shown in \cite{nightingale2021synthetic}, human accuracy for GAN-face detection is around 50\%$\sim$60\%, which shows that topics on improving or accommodating human performance are essential. We provide a comprehensive discussion in $\S$~\ref{sec:human:performance} on the topic of human visual performance for GAN-face detection.

The datasets are the driving force behind the rapid development of GAN models and GAN-face detection methods. We survey popular datasets and major evaluation metrics in $\S$~\ref{sec:datasets:eval}. 
For completeness, we also list other related surveys in $\S$~\ref{relatedsurvey}. 
In the foreseeable future, there are a number of critical problems that are yet to be resolved for existing GAN-face detection methods. With the development of the GAN models, it is thus important to anticipate such new developments and improve the detection methods accordingly. We discuss future research opportunities in $\S$~\ref{sec:future}. 

The contribution of this paper is summarized in the following:
\begin{itemize}
    \item To the best of our knowledge, this work is the first comprehensive review that discusses different types of GAN-face detection methods. We particularly include the explainable methods that provide interpretability of the decision process and results that ease human understanding.
    \item We organize and summarize the vast literature on GAN-face detection into four categories: 
    (1) deep learning-based methods, 
    (2) physical-based methods,
    (3) physiological-based methods, 
    and (4) human visual performance.

    \item  Human visual performance of recognizing GAN-faces is important, especially for people to check for their social networking and possible security or privacy violations. We provide a comprehensive discussion on human visual performance and strategies for checking against fake GAN-generated faces.
    
    \item We propose several issues associated with existing state-of-the-art methods and discuss future research directions.
    
\end{itemize}

\section{GAN Generation of Highly Realistic Faces}
\label{sec:GAN:face:gen}

We next provide a brief summary of mainstream methods for generating high-quality faces that most GAN-face detection works are targeting. Further details on the various kinds of GANs can be found in the surveys of \cite{jabbar2020survey,xia2021gan}.

In the past five years, numerous GAN models ({\em e.g.}, PG-GAN \cite{karras2017progressive}, BigGAN \cite{brock2018large}, StyleGAN \cite{karras2019style}, StyleGAN2 \cite{karras2020analyzing}, etc.) have been developed to synthesis and create realistic-looking face images with diversity from random noise input.
These GANs can effectively encode rich semantic information in the intermediate features \cite{bau2019semantic} and latent space \cite{goetschalckx2019ganalyze,jahanian2019steerability,shen2020interpreting} for high-quality face image generation. 
Moreover, these GANs can generate fake face images with various attributes, including various ages, expressions, backgrounds, and viewing angles. However, due to the lack of inference functions or encoders in GANs, such manipulations in latent space are only applicable to images generated from GANs, not to any given real images.

To address the above issue, GAN inversion methods can invert a given image back into its latent space of a pre-trained GAN model \cite{xia2021gan}. 
The GAN generator can then reconstruct the image accurately from the inverted code in approximation. This inversion method plays a key role in bridging real and fake face image domains. 
Therefore, it can significantly improve the quality of the generated face images and be applied widely in state-of-the-art GAN models including StyleGAN2 \cite{karras2020analyzing}, StyleGAN3 \cite{karras2021alias}, InterFaceGAN \cite{shen2020interpreting}, and Image2StyleGAN++ \cite{abdal2020image2stylegan++}.         


\section{GAN-face Detection Methods}
\label{method}

We organize existing GAN-face detection literature into four categories. Although there exist similarities of various methods {\em e.g.} across categories, we organize them primarily by their motivations and key ideas.
Table~\ref{ganfacedataset} summarizes mainstream GAN-face detection methods with the datasets used and performance comparison.

\subsection{Deep Learning-based Methods}
\label{sec:DL:methods}

Deep learning-based GAN-face detection methods extract signal-level features to train Deep Neural Network (DNN) classifiers to distinguish fake faces from real ones in an end-to-end learning framework \cite{jeong2022frepgan,fu2022detecting,cao2022end}.

\begin{figure}[t]
    \centering
    \includegraphics[width=.45\textwidth]{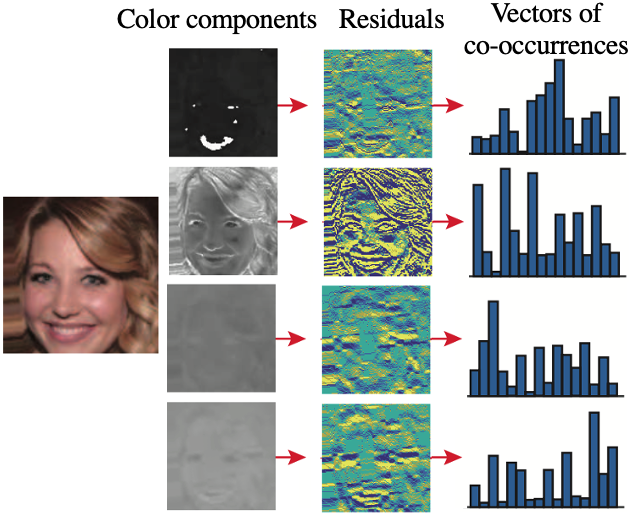}
    \vspace{-1em}
    \caption{\em The features of capturing color image statistics for training the classifier \cite{li2020identification}. }
    \vspace{-0.6em}
    \label{fig_colorfea}
\end{figure}

The earliest work of \cite{do2018forensics} employed VGG-Net \cite{simonyan2014very} for GAN-face detection. To train the network, real faces are collected from the CelebA face dataset \cite{liu2015faceattributes}, and fake faces are generated using DC-GANs \cite{radford2015unsupervised} and PG-GAN \cite{karras2017progressive}, where the VGG-16 architecture is used with pre-train weights of VGG-Face \cite{cao2018vggface2}. 
\cite{mo2018fake} found that signals in the residual field can serve as effective features to distinguish real and GAN-faces. They first processed the input faces with high-pass filters, and the resulting residuals were fed into deep networks for GAN-face detection.
\cite{li2020identification} identified GAN-faces by analyzing the chrominance color components. They first extracted a feature set to capture color image statistics (See Figure \ref{fig_colorfea}),
then use the concatenated features to train a GAN-face classifier. Similarly, \cite{chen2021robust} found that both the luminance and chrominance cues are useful for improving GAN-face detection.
More recently, \cite{fu2019robust} used a dual-channel CNN to reduce the impact of many widely-used image post-processing operations. The deep CNN of their network extracts features of the pre-processed images, and the shallow CNN extracts features from the high-frequency components of the original image.

{\bf GAN-face detection in real-world scenarios.}  The work \cite{hulzebosch2020detecting} developed a framework for evaluating detection methods under cross-model, cross-data, and post-processing evaluations, to examine features produced from commonly-used image pre-processing methods. More recently, many variants of feature-based models have been studied \cite{wang2020cnn,goebel2020detection,liu2020global,chen2022distinguishing}. 
However, the detection results from all these feature-based methods are not explainable, so it is unclear why the decision was given to any input face. 


{\bf One-shot, incremental and advanced learning.}
A one-shot GAN-face detection method was studied recently in \cite{mansourifar2020one}. 
Scene understanding is applied to determine out-of-context objects that appeared in the GAN-faces to distinguish GAN-faces from the real ones. 
The work \cite{marra2019incremental} applied incremental learning for GAN-faces image detection, where the key idea is to detect and classify new GAN-generated faces without decreasing the performance on existing ones.

\begin{figure}[t]
    \centering
    \includegraphics[width=.48\textwidth]{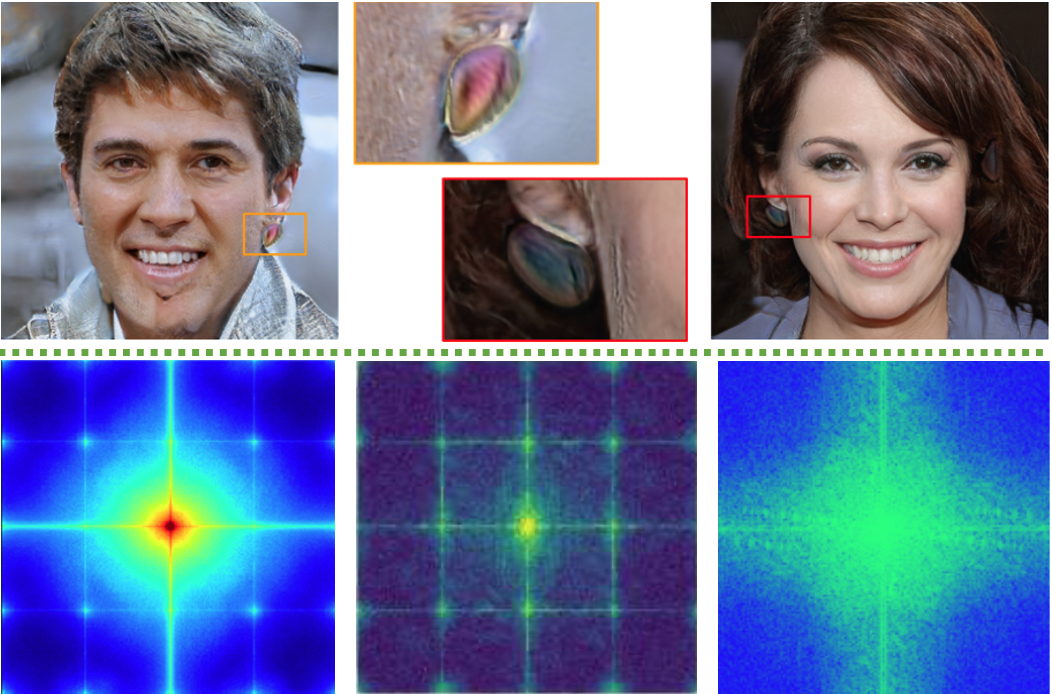}
    \vspace{-2em}
    \caption{\em \textbf{Top:} Visible color artifacts of GAN image \cite{gragnaniello2021gan}. \textbf{Bottom:}  Invisible artifacts of GAN image, averaged Fourier spectrum \cite{gragnaniello2021gan}, frequency analysis \cite{wang2020cnn}, frequency spectrum \cite{dong2022think}.}
    \vspace{-0.1em}
    \label{dlfeature}
\end{figure}

{\bf Difficulty Analysis.}
More difficulty analysis and systemic evaluations using state-of-the-art DNNs for GAN-face detection are investigated in \cite{gragnaniello2021gan,wang2019fakespotter,wang2020cnn,jeon2020t,dong2022think}, both visible and invisible artifacts are analyzed in these works (See Figure \ref{dlfeature}).
For example, \cite{wang2020cnn} find that the CNN-generated images share some common systematic flaws, resulting in them being surprisingly easy to spot for now. 
To investigate \textit{ Are GAN-generated images easy to detect?} \cite{gragnaniello2021gan} conducted the study to analyze the performance of the existing GAN-faces detection methods on different datasets and using different metrics. 
On the country, they concluded that we are still very far from
having reliable tools for GAN image detection.




Unfortunately, all aforementioned methods in this subsection can not provide explainable results. To overcome this shortcoming, an attention-based method was proposed in \cite{guo2021robust} to spot GAN-generated faces by analyzing eye inconsistencies. Specifically, this model learned to identify inconsistent eye components by localizing and comparing the iris artifacts. 
Visual results from \cite{guo2021robust} showed a clear difference between the attention maps of the irises from the GAN-faces and real ones. For GAN-faces the attention map highlighted the artifact regions on the irises, and for real faces, there is no significant concentration of the attention map. 
However, the attention map still cannot provide enough explainability to understand the behavior of the learned model.


In summary, Deep Learning-based methods achieved impressive performance on GAN-face detection \cite{gu2022exploiting}. However, it is difficult to explain or interpret the decision process of the learned model as a black box. 
Nonetheless, fake face detection in the real-world favors explainability, alongside from the overall accuracy.
Particularly, people do care more for use cases such as {\em ``This picture looks like someone I know, and if the AI algorithm tells it is fake or real, then what is the reasoning and should I trust?''}






\subsection{Physical-based Methods}
\label{sec:physical}

Physical-based methods identify GAN-faces by looking for artifacts or inconsistencies among the face and the physical world, such as the illumination and reflections in perspective.

\begin{figure}[t]
    \centering
    \includegraphics[width=.48\textwidth]{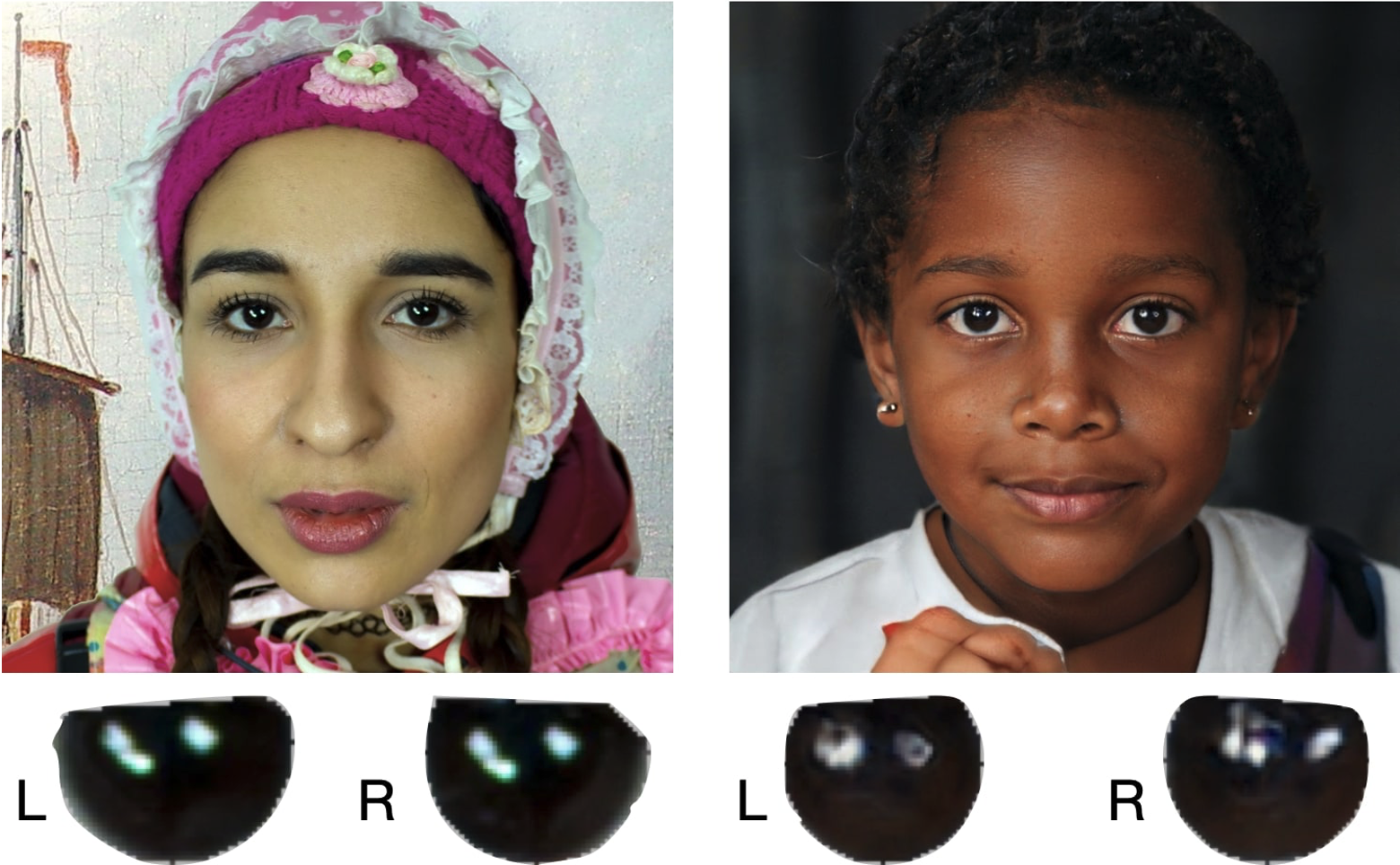}
    \vspace{-2em}
    \caption{\em  \textbf{Top:} Corneal specular highlights for a real human face (\textbf{left}) and a GAN-face (\textbf{right}) \cite{hu2021exposing}. \textbf{Bottom:} The corneal regions are isolated and scaled for better visibility. Note that the corneal specular highlights for the real face have strong similarities while those for the GAN-faces are different.}
    \vspace{-0.1em}
    \label{fig_hu2021exposing}
\end{figure}

The early work of 
\cite{johnson06specular} analyzed the internal camera parameters and light source directions from the perspective distortion of the specular highlights of the eyes to 
reveal traces of image tampering.
Recently, \cite{matern2019exploiting} identified early versions of GAN-faces \cite{karras2017progressive} based on an observation that the specular reflection in the eyes of GAN-faces is either missing or appearing as a simple white blob.
However, such artifacts have been largely corrected in recent GAN-faces such as StyleGAN2.
The method of \cite{hu2021exposing} looked for inconsistency between the two eyes to identify GAN-generated faces. Specifically, the corneal specular highlights of the eyes are detected and aligned for pixel-wise Intersection of Union (IoU) comparison. 
As shown in Figure \ref{fig_hu2021exposing}, the assumption is that real human eyes captured by a camera under a portrait setting should exhibit a strong resemblance between the corneal specular highlights between the two eyes. In contrast, this assumption is not true for GAN synthesized eyes, where inconsistencies include different numbers, different geometric shapes, or different relative locations of the specular highlights. 
However, this method operates on strong assumptions of the frontal portrait pose, far away lighting source(s), and the existence of the eye specular highlights. When these assumptions are violated, false positives may increase significantly. 

In summary, the physical-based detection methods are more robust to adversarial attacks, and the predicted results afford intuitive interpretations to human users~\cite{hu2021exposing}.

\subsection{Physiological-based Methods}
\label{sec:physiological}

Physiologically-based methods investigate the semantic aspect of the human faces \cite{ciftci2020fakecatcher}, including cues such as symmetry, iris color, pupil shapes, {\em etc.}, where the identified artifacts are used for exposing GAN-faces. 

Early works of \cite{marra2019gans,yu2019attributing,mccloskey2018detecting} indicated that StyleGAN \cite{karras2017progressive} generated faces contain obvious artifacts including asymmetric faces and inconsistent iris colors \cite{matern2019exploiting}.
\cite{yang2019exposinggan}
found that GAN can generate facial parts ({\em e.g.}, eyes, nose, skin, mouth) with a great level of realistic details, yet there is no explicit constraint over the locations of these parts on the face. In other words, the facial parts of GAN-faces may not appear to be coherent or natural-looking, when compared to real faces.
They indicated that these abnormalities in the configuration of facial parts in GAN-faces could be revealed using the locations of the facial landmark points ({\em e.g.}, tips of the eyes, nose, and mouth), which can be effectively detected using automatic algorithms. The normalized locations of these facial landmarks can be used features to train a classifier to identify GAN-faces. 
However, GAN-face generation has also improved on the other hand. Face images generated by StyleGAN2 have improved greatly in quality and are free of obvious physiological artifacts \cite{karras2017progressive,karras2019style,karras2020analyzing}. 
And the synthesis process of GAN-faces is further optimized in StyleGAN3. It exhibits a more natural transformation hierarchy of different scales of features. They are fully equivariant to translation and rotation, which further improved the physiological consistency of the generated faces.



\begin{figure}[t]
    \centering
    \includegraphics[width=\linewidth]{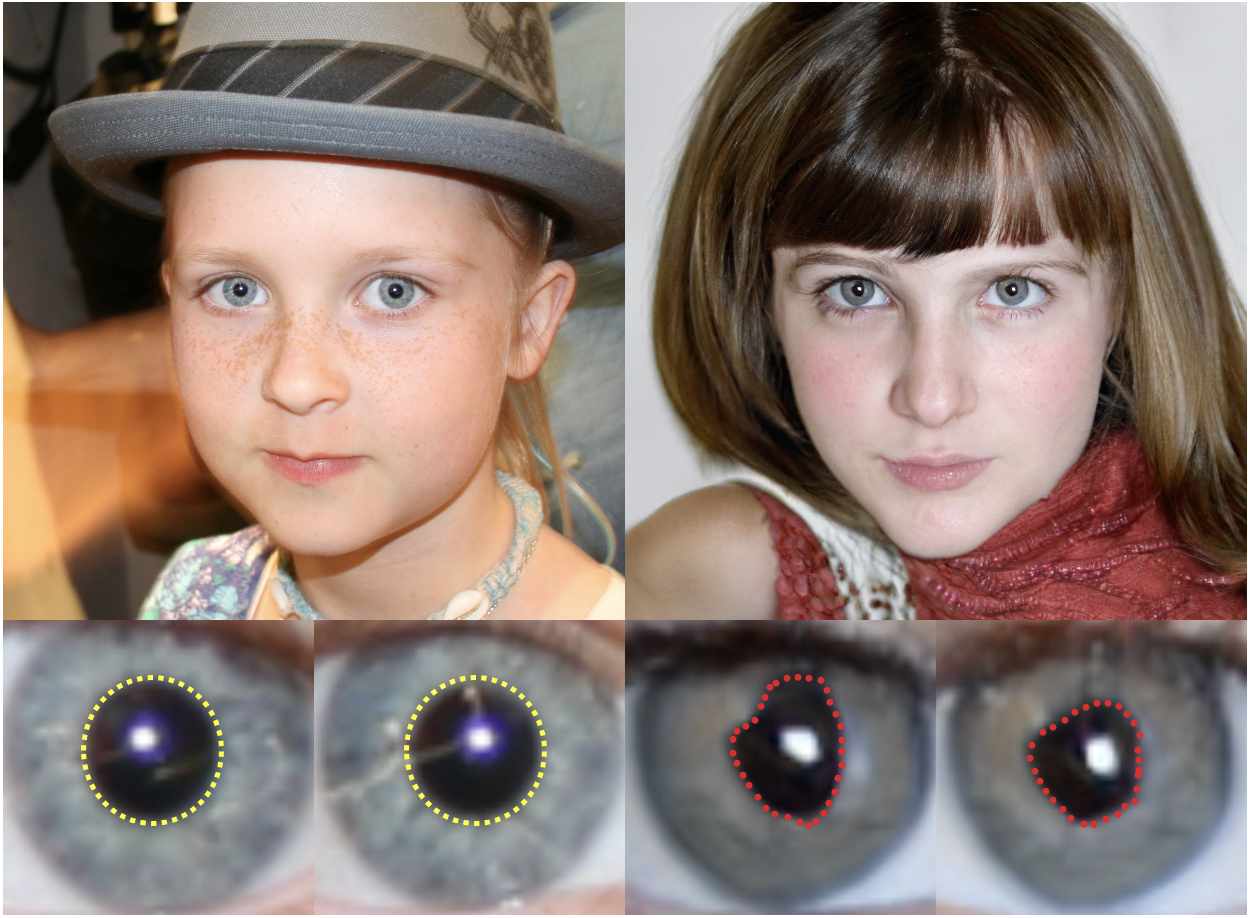}
    \label{figpupilshape}
    \vspace{-2em}
    \caption{\em Pupils of real (\textbf{left}) human face and GAN-face (\textbf{right}) \cite{guo2021eyes}. Note that the pupils for the real eyes have strong circular shapes ({yellow}) while those for the GAN-generated pupils are with irregular shapes ({red}). }
    \label{figpupilshape}
\end{figure}

A relatively new physiological-based GAN-face detection method is proposed in \cite{guo2021eyes}, motivated by a simple observation that GAN-faces exhibit a common artifact of irregular pupil shapes. Specifically, pupils from real human faces should appear to be a smooth circle or ellipse; in contrast, pupils from GAN-faces can appear with irregular shapes or boundaries (See Figure \ref{figpupilshape}). 
This artifact is universal for all known GAN models up to date (including PG-GAN~\cite{karras2017progressive}, StyleGAN3~\cite{karras2021alias}, and SofGAN~\cite{chen2021sofgan}), and this artifact occurs in eyes from the synthesized humans and animals.
One fundamental reason for the existence of such artifacts in GAN-generated faces is due to their lack of understanding of human eye anatomy, particularly the geometry and shape of the pupils. 
The method of \cite{guo2021eyes} first localize the eyes and segment out the pupil region. Next, an ellipse model is parametrically fit to the pupil boundary. Boundary IoU \cite{cheng2021boundary} is then calculated between the extracted pupil mask and the fitted ellipse to estimate the ``circularness'' of the pupils.
However, false positive can arise in rare cases of non-elliptical pupils in real faces due to diseased or infected eyes.  


In summary, physiological-based method comes with stronger interpretability. However, like other forensic approaches, environmental constraints such as occlusion and visibility of the eye from the face image is still a major limitation. 
\begin{table*}[t]
\centering
\scalebox{0.84}{
\begin{tabular}{|c|c|c|c|c|c|c|}
\hline
\multicolumn{1}{|c|}{\textbf{Paper}} & \textbf{Category}     &\textbf{Method}  & \multicolumn{1}{c|}{\textbf{Real Face ($\#$test)}} & \multicolumn{1}{c|}{\textbf{GAN Face ($\#$test)} } & \multicolumn{1}{c|}{ \textbf{Performance} }  \\ 
\hline
\hline  
    \cite{nightingale2021synthetic}             & Human      &       Visual  & FFHQ (400)                              & StyleGAN2 (400)             & Acc: 0.5$\sim$0.6                        \\ \hline
       \cite{lago2021more}             & Human  &            Visual  & FFHQ (150)                              &  StyleGAN2, etc. (150)             & Acc: 0.26$\sim$0.8                         \\ \hline
       
   \cite{do2018forensics}                        & DL  &     CNN      & CelebA (200)                 & PGGAN, DCGAN (200)                   & Acc: 0.80                                          \\ \hline

       \cite{dang2018deep}                        & DL  &     CNN     &         CelebA (1250)               &          PCGAN (1250)   & Acc: 0.98                                        \\ \hline

       \cite{mo2018fake}                       & DL  &       CNN       &  CelebA-HQ (15K)                            & PGGAN (15K)              &             Acc:  0.99                    \\ \hline
       
       \cite{nataraj2019detecting}                        & DL  &      CNN & CelebA (500)                      & StarGAN (4498)                  & Acc:0.99   
                  \\ \hline
          
          \cite{fu2019robust}                        & DL  &      CNN &       CelebA-HQ (7K)       &    PGGAN (7K)         &    Acc: 0.98
                 \\ \hline

          \cite{marra2019incremental}                        & DL        &  Incremental Classifier                        & -    & StarGAN (2.4K), etc.
          & Acc: 0.815$\sim$1       \\ \hline

       \cite{mansourifar2020one}                        & DL       &    Out of context object detection      & -  &        StyleGAN (100)           &        Acc: 0.80                                 \\ \hline
       
       \cite{wang2019fakespotter}                        & DL       &   DNN      & FFHQ (1K)                      & StyleGAN2 (1K), etc.                  & Acc: 0.88$\sim$0.991                                         \\ \hline
       
       \cite{li2020identification}                        & DL       & Disparities in Color Components     & CelebA-HQ, FFHQ (50K)      & StyGAN, ProGAN (50K)   &                    Acc: 0.997                      \\ \hline
       
       \cite{wang2020cnn}                        & DL  &       CNN        & FFHQ  (1K)                      & StyleGAN (1K)                  & Acc: 0.84                                        \\ \hline
        \cite{hulzebosch2020detecting}                        & DL        &  ForensicTransfer        &             FFHQ (3K), etc.       &  StyleGAN (3K), ProGAN (3K), etc. & Acc:   0.01$\sim$1                                \\ \hline
        \cite{goebel2020detection}                        & DL        &  CNN       &   CelebA (164), CelebA-HQ (1.5K)                 & StarGAN (1476), ProGAN (3.7K)  & Acc: 0.6768$\sim$0.849                                     \\ \hline
        \cite{liu2020global}                        & DL        &   CNN      &        FFHQ (10K), CelebA-HQ (10K)             &  StyleGAN (10K), PGGAN (10K), etc. & Acc: 0.9854$\sim$0.991                                    \\ \hline
       
       \cite{chen2021locally}                       & DL       &  Xception       &  FFHQ (7K)                            & LGGF (14K)              & Acc: 0.99                                 \\ \hline
       \cite{chen2021robust}                       & DL       &  Improved Xception       &                      CelebA (202,60)    &  PGGAN (202,60)            &  Acc:  0.713$\sim$0.977                            \\ \hline   
        
    \cite{gragnaniello2021gan}                        & DL        &   CNN      &       RAISE ($\leq$7.8K)             &  StyleGAN2 (3K), ProGAN (3K), etc.  & Acc: 0.928$\sim$0.999                                     \\ \hline
    \cite{chen2022distinguishing}                        & DL        &    CNN     &     FFHQ (20K)  & StyleGAN (20K), etc.   & Acc: 0.9895$\sim$1                                    \\ \hline
      
    \cite{guo2021robust}                       & DL       &  Residual Attention       & FFHQ (748)                               & StyleGAN2 (750)                        & AUC: 1                                        \\ \hline  
    \cite{nowroozi2022detecting} & DL       &   Cross-Co-Net       & FFHQ   (4K)                  &  StyleGAN2 (4K)                        & Acc: 0.998                                     \\ \hline   
    \cite{dong2022think} & DL       &   Enhanced spectrum based CNN       & FFHQ   (3.2K)                  &  StyleGAN, StyleGAN2 (3.2K)                        & Acc: 0.95 $\sim$ 0.96                                      \\ \hline  
      \cite{wangeyes}                       & DL       &  Siamese  Network       & FFHQ, CelebA-HQ (10K)                              & ProGAN, StyleGAN3 (20K)                        & AUC: 0.996 $\sim$ 1                                       \\ \hline      
   \cellcolor{gray!20}   \cite{hu2021exposing}                       & Physic       &   Corneal specular highlight       & FFHQ (500)                               & StyleGAN2 (500)                        & AUC: 0.94                                        \\ \hline
       \cellcolor{gray!20}   \cite{matern2019exploiting}                       & Physiology     &   Eye color      & CelebA (1K)                            & ProGAN (1K), Glow (1K)                & AUC: 0.70$\sim$0.85                                    \\ \hline

     \cite{yang2019exposinggan}                       & Physiology     &     Landmark locations    &   CelebA ($\geq$50K)                          &        PGGAN (25K)         &  AUC: 0.9121$\sim$0.9413                           \\ \hline
    
    \cellcolor{gray!20}  \cite{guo2021eyes}                       & Physiology       &   Irregular pupil shape       & FFHQ (1.6K)                            & StyleGAN2 (1.6K)               & AUC: 0.91                                   \\ \hline
\end{tabular}
}
\vspace{-0.1em}
\caption{\em Summary of GAN-face detection methods with the corresponding datasets, statistics and performance scores. The gray rows highlight those where {\em individual} predicted results of the method are {\bf explainable} to humans. Note that datasets used in the works are self-collected and can contain different subsets across papers. So the performance scores do not represent fair comparisons.
}
\label{ganfacedataset}
\vspace{-0.1em}
\end{table*}

\subsection{Human Visual Performance}
\label{sec:human:performance}

Although many automatic GAN-face detection algorithms have been developed, human visual performance in identifying and exposing GAN-faces has not been investigated sufficiently. Compared with other AI problems such as image recognition, GAN-face detection is a much more challenging problem for human eyes. Thus, it is important to study how well human eyes can identify GAN-faces and the related social impacts and ethical issues \cite{barrington2023comparative}.


Standard metrics for evaluating the effectiveness of automatic algorithms in detecting GAN-faces include ROC analysis and Precision-Recall. While these metrics can be applied to study human perceptual performance, they are not directly suitable in reflecting the true deceptiveness of the highly realistic GAN-faces for the general public. Human performance is largely biased, and with weak but proper hints (such as looking for the correct physiological cues), human performance in identifying fake faces can boost greatly \cite{khodabakhsh2019subjective}.

An early work \cite{lago2021more} conducted a study to measure the human ability to recognize fake faces. Their dataset consists of 150 real faces and 150 GAN faces. Real faces are selected from the Flickr-Faces-HQ (FFHQ) dataset, and GAN-Faces are generated from state-of-the-art GANs, including PG-GAN, StyleGAN, and StyleGAN2. The 630 participants sequentially completed 34 tasks to distinguish 30 faces each time. Those faces were randomly selected in equal portions from each category. Results showed that participants had lost the ability to judge newer GAN-faces. Accuracy is not impacted when the test speeds up or the participants have seen similar synthetic faces produced by the generators before.

\begin{figure}[t]
    \centering
    \includegraphics[width=.45\textwidth]{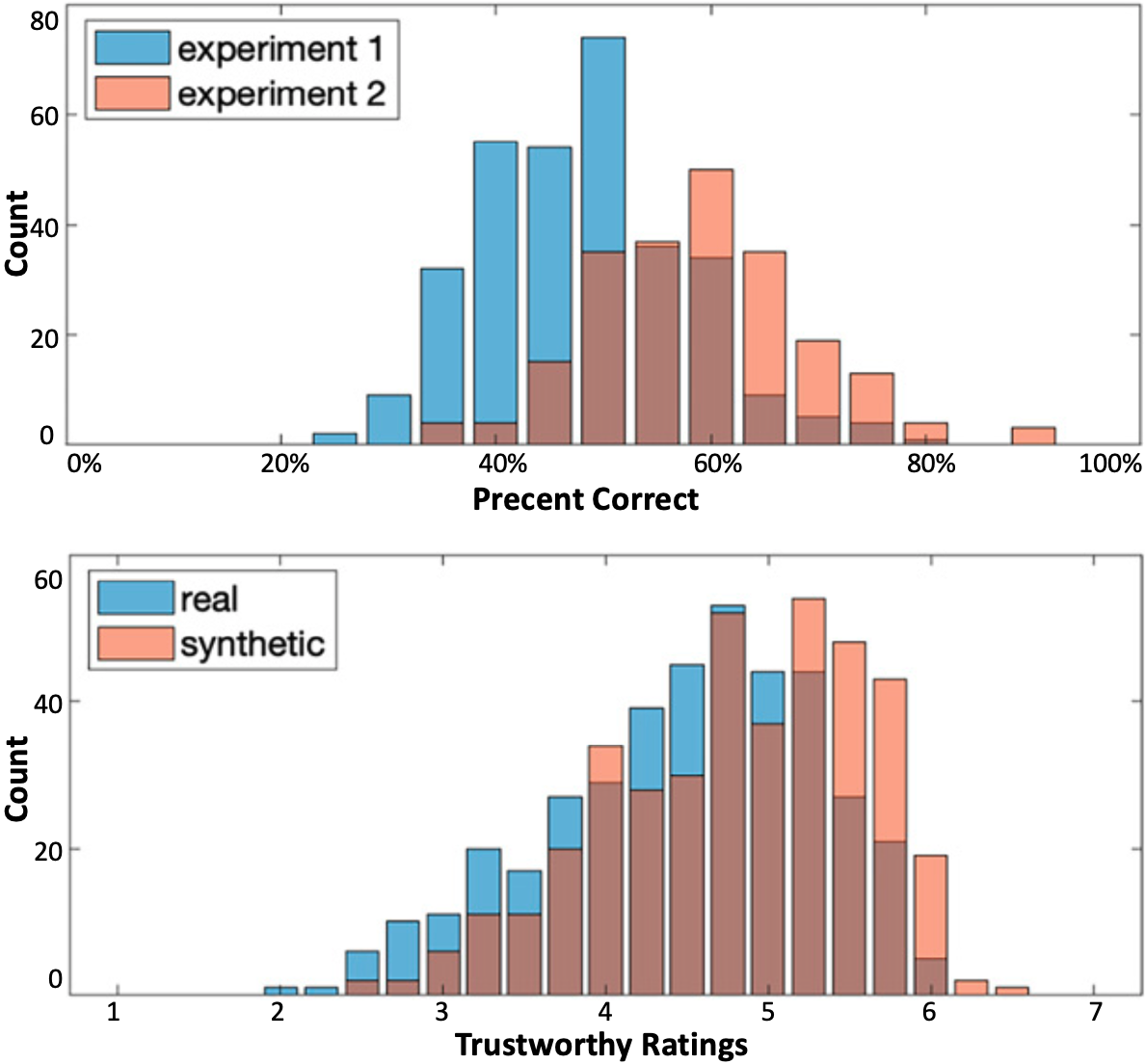}
    \caption{\em Human visual performance \cite{nightingale2022ai}. \textbf{Top:} Average performance of experiments 1 and 2, the accuracy is around 50$\%$. In experiment 2, the training and feedback improves average performance a little bit.  \textbf{Bottom:} Trustworthy ratings for experiment 3, a rating of 1 corresponds to the lowest trust.}
    \vspace{-0.1em}
    \label{fig_vss} 
\end{figure}

A recent work \cite{nightingale2021synthetic} examined people's ability to discriminate GAN-faces from real faces.
Specifically, 400 StyleGAN2 faces and 400 real faces from the FFHQ dataset are selected with large diversity across the genders, ages, races, {\em etc.}, and two sets of experiments are conducted. 
In the first set of experiments, 315 participants were shown a few examples of GAN-faces and real faces, and around $ 50\%$ of accuracy is obtained. In the second set of experiments, 170 new participants were given a tutorial consisting of examples of specific artifacts in the GAN-faces. Participants were also given feedback afterward. 
However, it was found that such training and feedback only improve a little bit of average accuracy. Therefore, this work concluded that the StyleGAN2 faces are realistic enough to fool both naive and trained human observers, more extended studies are summarized in \cite{nightingale2022ai}, the experiment 3 is conducted to further investigate whether synthetic faces activate the same judgements of trustworthiness. A perception of trustworthiness could also help distinguish real from GAN-faces. Their experimental results are shown in Figure \ref{fig_vss}.
However, no information on what synthesis artifacts are provided for participant training in this study. 
We believe there is still space to improve human capability in discerning GAN-faces if sufficient hints are provided, including physiological cues ({\em e.g.} pupil shapes \cite{guo2021eyes}) and dataset statistics ({\em e.g.} GAN-faces are usually trained with FFHQ samples that are biased toward portrait faces and celebrity styles).

GANs are under active development, so it is expected that the difficulty of discerning GAN-faces will continue to increase.
It is important to find generic and consistent cues for human eyes to effectively distinguish GAN-faces. Typically, useful cues are generally universal for exposing other types of AI tampered faces, including morphed faces, swapped faces, painting faces. 
Recently, an open platform to study whether a human can distinguish AI-synthesized faces from real faces visually by using the cues is developed in \cite{guo2022open}.
The discovery of such cues can also be leveraged for improving the GAN face synthesis algorithm to produce faces that are even harder to distinguish for human eyes.

In summary, there is no doubt that the studies of human visual performance are invaluable to research detection techniques as well as a better understanding of the insufficient of the GAN-faces.

\section{Datasets and Performance Evaluation}
\label{sec:datasets:eval}

With the rapid development of AI discriminative and generative models, many human facial datasets have been constructed. Among these datasets, real face images are mainly collected from the FFHQ dataset \cite{karras2019style}, CelebA \cite{liu2015faceattributes}, CelebA-HQ \cite{karras2017progressive}, RAISE~\cite{dang2015raise} {\em etc.}. Synthesized face images are collected using state-of-the-art GAN models and LGGF~\cite{chen2021locally}.

Early GAN-faces datasets are mainly comprised of PGGAN, and recent datasets are typically based on StyleGAN2.
NVIDIA has recently curated a StyleGAN3 generated set at \url{https://github.com/NVlabs/stylegan3-detector} that can be used to evaluate GAN-face detection performance.
Table~\ref{ganfacedataset} list mainstream datasets for GAN-face detection. 
Note that datasets used for each work are self-collected and can contain different subsets across papers.
This is due to that only specific subsets are relevant to individual methods. For example, in \cite{guo2021eyes}, 
only face images with visible eye pupils are used for training and evaluation.

As GAN-face detection is a binary classification problem, {\bf evaluation metrics} are typically based on Accuracy, Precision-Recall, ROC analysis, and AUC.
To the best of our knowledge, a sufficiently large-scale benchmark dataset for empirical evaluation of GAN-face detection is still lacking.

\section{Related Surveys}
\label{relatedsurvey}



Although the scope of this survey is on the detection of \textcolor{black}{\textit{GAN-based entire face synthesis}}, the GAN-face detection task is closely related to other fake face detection tasks \cite{ni2022core,fei2022learning}. 
For completeness, we also discuss other surveys in the related fields. 

\noindent \textbf{GAN-face Detection}. A related survey in \cite{liu2020survey} only discussed DL-based GAN-face detection works and ignored other significant non-DL-based works. Their survey neglects the interpretability issues, which is crucial for applying DL-based methods for detecting GAN-faces in practice.


\noindent \textbf{Morphed-face Detection}. Morphed-face detection aims to detect images that have been merged by two or more face images \cite{pikoulis2021face}. 
Morphed-face detection is a challenging problem due to the complex nature of the morphing techniques used to create these images. Therefore, the development of effective and accurate morphed-face detection methods is of utmost importance to prevent potential harm caused by these fraudulent activities \cite{scherhag2019face}. 
The current surveys provide an overview of the recent advances towards both the generation and the detection of morphing face \cite{saleem2023techniques, rajaface}. 
However, the motivation for generating morphed faces is usually different from GAN-face, these morphed faces are often created for identity theft and profile impersonation.

\noindent \textbf{Manipulated-face Detection}. Manipulated-face detection involves the identification of images that have been manipulated or tampered with by AI algorithms \cite{pashine2021deep}. 
The increasing availability of AI algorithms has made it easier to create manipulated images for fraudulent purposes, such as face swapping \cite{jiang2022facke}, and facial manipulation \cite{liu2022gan}.
The current surveys \cite{nickabadi2022comprehensive,tolosana2020deepfakes} indicate that the development of reliable and robust manipulated-face detection methods is essential to safeguard the authenticity and integrity of digital media and prevent harm caused by the misuse of manipulated images. 

As previously mentioned, it's important to note that face manipulation and GAN-based face generation are distinct techniques. While GANs generate faces from scratch, face manipulation involves altering or modifying existing face images using various techniques.

\begin{figure}[t]
\centering
\includegraphics[width=.45\textwidth]{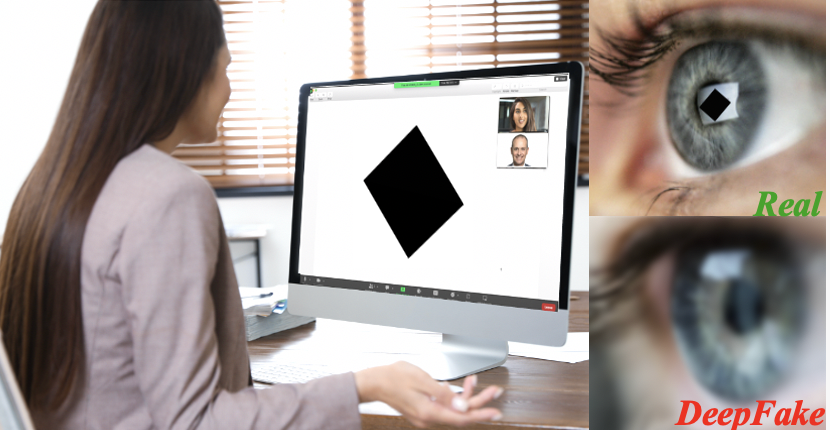}
 \vspace{-1em}
\caption{\em Video conferencing DeepFakes detection \cite{guo2022detection}. \textbf{Left:} A video call attendant is being actively authenticated with the live patterns shown on the screen.  \textbf{Right:} A real person's cornea will produce an image of the pattern shown on the screen while a real-time DeepFake cannot. }
\label{fig_zoom}
\vspace{-1em}
\end{figure}

\noindent \textbf{DeepFakes Detection}. DeepFakes detection involves detecting and identifying images, videos, audio, and text that have been generated or manipulated using artificial intelligence techniques \cite{nguyen2019deep,yu2021survey,shiohara2022detecting}. 
DeepFakes are often created to deceive and manipulate viewers by inserting fake information into real events or spreading misinformation or creating fake news \cite{swathi2021deepfake,passos2022review,bhagtani2022overview}. 
The use of DeepFakes poses a significant threat to the authenticity and credibility of social media \cite{bohavcek2022protecting}, making it essential to develop reliable and effective DeepFakes detection methods \cite{mirsky2021creation,deb2023unified,juefei2022countering}.

The DeepFakes detection is primarily focused on identifying and combating the spread of fake news and misinformation \cite{zhang2022deepfake,malik2022deepfake,zobaed2021deepfakes,le2022robust}. 
The current surveys can help prevent the potential harm caused by the misuse of DeepFakes and ensure that the information presented is accurate and trustworthy \cite{tolosana2020deepfakes,juefei2021countering}.
Furthermore, other surveys have been conducted with a different perspective compared to existing survey papers, for example, the survey \cite{khanjani2021deep} mostly focuses on the audio deepfakes that are overlooked in most of the previous surveys, \cite{masood2023deepfakes} focus on the audio-visual DeepFakes generation and detection, and the benchmarks \cite{altuncu2022deepfake,lin2022towards,pham2021dual,li2023continual}.

Moreover, the COVID pandemic \cite{li2020using} has led to the wide adoption of online video calls. The increasing reliance on video calls provides opportunities for new impersonation attacks by fraudsters using the advanced real-time DeepFakes \cite{Gerstner_2022_CVPR}. 
Video conferencing DeepFakes poses new challenges to detection methods, which have to run in real-time as a video call is ongoing. More recently, the video conferencing DeepFakes detection methods (Figure \ref{fig_zoom}) are also developed \cite{guo2022detection}, the topic of video conferencing DeepFakes detection has not been covered in the previous surveys.

In summary, DeepFakes detection is a broad topic. Although GAN-face detection is often included in related surveys, it may lack a detailed analysis of the methods used to detect GAN-generated faces. 
The in-depth analysis and discussion of the GAN-face detection methods in our survey, such as analyzing the explainable features that differentiate real and fake faces, can help researchers develop more effective detection techniques. 
By further investigating the GAN-face detection methods, we can better understand the challenges and limitations of DeepFakes detection and develop more accurate and reliable methods for detecting GAN-based DeepFakes.


\section{Future Directions}
\label{sec:future}

After reviewing existing methods of GAN-face detection with identified advantages and limitations, we next discuss future research directions that are promising for developing forensic algorithms that will be more effective, interpretable, robust, and extensible.

\subsection{Against the Evolution of GAN models}

Although the existing GAN models can not generate perfect fake faces due to known vulnerabilities, more powerful GAN models are under active development and certainly will come out in the near future. 
We anticipate that the known artifacts of GAN-faces ({\em e.g.} inconsistent corneal specular highlights \cite{hu2021exposing}, irregular pupil shapes \cite{guo2021eyes}, symmetry inconsistencies such as different earrings, {\em etc.}) can be fixed by incorporating relevant constraints to existing GAN models; however, how best to effectively enforce such constraints are still open questions. 
More powerful deep neural network architectures, training tricks, and larger training data will continue to push the state-of-the-art GAN models.
For example, StyleGAN3 \cite{karras2021alias} presents a comprehensive overhaul of all signal processing aspects of StyleGAN2 to improve the texture and 3D modeling of the GAN-generated faces.
The demands for searching for effective cues for exposing new GAN-faces and developing more powerful GAN-face detection methods continue to rise.

{\bf Low-power demands.}
In addition, computationally effective GAN-face detectors that can run on edge devices are of practical importance. Since GAN-faces can directly cause concerns and impacts regarding identities and social networks, forensic analytics should ideally be able to run on smartphones. Research on how best to migrate high FLOPS GPU models toward mobile applications has practical needs.


\subsection{How to Develop Good Interpretation Methods} 

One critical disadvantage of many GAN-face detection methods is that they do not afford interpretability for the predicted results. Methods based on the widely-used attention mechanism \cite{guo2021robust} can not provide an interpretable explanation of the prediction results. Although the attention heat map highlights pixels that the network predicts, the mechanism can not tell {\em why} these pixels are selected that improves performance.
Furthermore, although the current physical \cite{hu2021exposing} and physiological-based methods \cite{guo2021eyes} can provide interpretability of their predicted results, their assumptions are {\em per-cue} based (such as the iris or pupil inconsistencies) that might not be extensible to future GAN models that are specifically designed. 
How best to develop an end-to-end mechanism that can effectively leverage physical and physiological cues for GAN-face detection is still an open research question. 






{\bf Learning multiple cues.}
From the numerous GAN-detection methods being surveyed, we observe that methods depending on a single cue or a few cues cannot retain performance, extensibility, and explainability at a time when dealing with complex real-world challenges such as occlusions and noisy data.
It is difficult for features drawn from a single cue to cover multiple characteristics or artifacts. So how best to improve the generalization of the learning system, and how best to integrate or fuse the learning of multiple cues into a unified framework will be the key. Ensemble learning \cite{sagi2018ensemble}, multi-model/task learning and knowledge distillation \cite{gou2021knowledge} are directions that future GAN-face detection models can benefit.



\subsection{Robust to Adversarial Attack}

As DNNs are widely used in GAN-face detection (either as a component or as the main model), DNNs are known to be vulnerable against {\em adversarial attacks}, which are based on intentionally designed perturbations or noises that are particularly effective and harmful to the DNNs.
With the increasing effectiveness of adversary attack technologies \cite{hu2021tkml}, research efforts start to focus on attacking fake face detectors particularly instead of focusing on general classifiers. Anti-forensics methods for evading fake detection via adversarial perturbations have been studied including \cite{carlini2020evading,gandhi2020adversarial}. 
These methods of attacking fake image detectors usually generate adversarial perturbations to perturb almost the entire image, which is redundant and can increase the perceptibility of perturbations. 
%
\cite{liao2021imperceptible} introduced a sparse attacking method called Key Region Attack to disrupt the fake image detection by determining key pixels to make the fake image detector only focus on these pixels.
Their adversarial perturbation appears only on key regions and is hard for humans to distinguish.
In general, future GAN-face detection methods need to be cautious in dealing with adversary attacks. 






\subsection{Imbalanced Distribution of Data}
\label{sec_Imbalanced}

In the real world, real faces usually significantly outnumber GAN-generated faces in online applications.
The data distribution for GAN-face detection is very imbalanced. Thus, the performance of GAN-face detection methods trained on balanced datasets may degrade when used for real-world applications, {\em e.g.} high accuracy but low sensitivity for spotting GAN-faces in practice.

As an initial effort, the method of \cite{guo2021robust,pu2022learning} 
addresses the imbalance learning issues by maximizing the ROC-AUC via 
approximation and relaxation of the AUC using Wilcoxon-Mann-Whitney (WMW) statistics.
Experimental results showed the robustness of the model learned from imbalanced data.
Looking forward, how best to deal with learning from extremely imbalanced data in real-world settings is an open question.


\subsection{Handling Mixtures with Other Fake Faces }

As face image tampering technologies continue to develop, including Face Morphing \cite{nightingale2021perceptual}, Face swapping \cite{perov2020deepfacelab}, Diffusion synthesized faces \cite{bohavcek2023geometric}, {\em etc.}, GAN-face detection forensics should be robust enough to deal with the mixture of face faking or synthesis methods.
In addition to the {\em detection} of GAN-faces, the {\em attribution} (find out what tools were used in the generation and the source where the faces come from) and {\em characterization} (find out the purpose of the generation and if the intention is malicious) are with growing importance. The DARPA Semantic Forensic (SemaFor) program \url{https://www.darpa.mil/program/semantic-forensics} of the U.S. is an ongoing effort that addresses these issues.



\section{Conclusion}

This paper presents a comprehensive, up-to-date review of GAN-face detection methods. We have reviewed the state-of-the-art models from multiple perspectives as well as provided details of major approaches. 
Although GAN-face detection has made notable progress recently, there is still significant room for improvement. 
Detecting GAN-faces in real-world settings remains challenging and with high demand, and we have discussed future research directions. 
The surveyed techniques and cues can also benefit the detection of other fake face-generation tools such as face morphing and swapping.

\ack This work is supported by the National Science Foundation, Project SaTC-2153112.


\bibliography{main}
\end{document}